\documentclass[11pt]{article}
\usepackage[margin=1in]{geometry}
\usepackage{graphicx,amsmath,amssymb,booktabs,hyperref,xcolor}
\usepackage[T1]{fontenc}
\usepackage{natbib}
\graphicspath{{fig/}}
\hypersetup{colorlinks=true,linkcolor=blue,citecolor=blue,urlcolor=blue}

\newcommand{\lca}{\lambda_{\mathrm{ca}}}
\newcommand{\dnorm}{D_{\mathrm{norm}}}
\newcommand{\Tstar}{T^{*}}

\title{What Iterated Self-Feeding Probes of Language Models Measure\\
{\large and a test that separates the construction from the model}}

\author{Nicol\'as Vera Z\'u\~niga\\
Independent Researcher, Chile\\
\texttt{nicovera@quetru.cl}
}
\date{}

\begin{document}
\maketitle

\begin{abstract}
A growing class of methods probes a language model by feeding it its own output: self-consistency,
iterated refinement, agentic loops. We ask what such a probe measures, in a construction chosen to
make the question sharp: a ring of token cells resampled in place by the model's own windowed
conditional $p_r(x_i \mid x_{i\pm r})$. The substrate is Glauber dynamics on token sequences and is
not new; what we change is the coupling. Advancing two rings that differ in one token under
\emph{common random numbers} makes undamaged copies diverge by exactly zero, so damage spreading
becomes measurable where a maximal coupling gives mixing times instead. The answer
is that it measures two different things at once, in readings that look alike. Some quantities are
fixed by the construction: the damage light cone is kinematic, and the radius scaling of the
token-space Lyapunov exponent $\lca(r)$ is \emph{model-invariant} across 19 models and two scale
ladders spanning $70\times$. Others genuinely track the model: $\lca$ crosses zero at a
reproducible point in training, and the attractor share ranks models consistently however the
lattice is built. Left
undistinguished, the first kind is readily mistaken for the second --- we did so ourselves for four
months, and report a phase transition we measured to three decimal places that belongs to the probe
rather than to any language model. We give the test that separates them: hold the construction
fixed and vary the model, or hold the model fixed and vary the construction, and see which readings
move. We validate the instrument by reproduction first, recovering a Domany--Kinzel damage field
\emph{bit-exactly} against an independent prediction, and we report the estimator failures that
this discipline caught --- four retracted verdicts, each on a quantity that looked like a
measurement. The methodology ships as a package.
\end{abstract}

\section{Introduction}
\label{sec:intro}

Feeding a language model its own output is now routine. Self-consistency samples a model repeatedly
and aggregates \citep{selfconsistency}, on reasoning traces the model produced itself
\citep{cot}; iterated refinement asks a model to revise its own answer against its own critique
\citep{selfrefine}; agentic loops return a model's output to its input for many turns. Each of these is a \emph{dynamical system} built from a
static predictor, and each invites a natural question: what do the dynamics tell us about the
model?

This paper answers a prior question, because we found it has an unobvious answer. Before asking
what an iterated probe reveals, one must ask what it \emph{measures} --- and the answer is that it
measures the model and the probe together, in quantities that are not distinguishable by
inspection.

We work in a construction chosen to make the confound as stark as possible: a ring of $N$ token
cells, each resampled in place from the model's own windowed conditional $p_r(x_i \mid x_{i\pm r})$
at temperature $T$, so that the model's output at every site is part of its own input at the next
step and no external text enters after initialisation.

\textbf{The substrate is not new, and we do not claim it.} In-place resampling of masked tokens is
Glauber dynamics on token sequences, and it has been studied as such \citep{glauber_mlm}. What we
change is the \emph{coupling} and therefore the observable. That work couples two chains maximally
and measures mixing time and metastability --- how long the chain takes to forget where it started.
We couple with \emph{common random numbers}: two rings, identical but for one flipped token,
advanced with the same stream of uniform variates and the same visit order. Maximal coupling and
CRN coupling are provably distinct constructions, and they answer different questions. Under CRN,
two rings whose windows agree draw the \emph{same} token, so undamaged copies diverge by exactly
zero and every difference is attributable to the injected flip. That is what makes damage
\emph{spreading} measurable at all: its exponential growth rate is a token-space Lyapunov exponent
$\lca$, its saturating level normalised by an independent-noise floor is a damping length $\dnorm$,
and its spatial extent is a light cone. None of these is a mixing time.

\paragraph{Contributions.}
\begin{enumerate}
\item \textbf{A validated instrument} (\S\ref{sec:validation}). We calibrate by
  \emph{reproduction}: on the Domany--Kinzel automaton, where the damage field is provably the
  automaton itself, an independent prediction matches ours bit-exactly --- zero mismatching cells,
  with a nonzero off-line control that must fail. The instrument also separates elementary CA rules
  by whether damage survives, and recovers known transition matrices.
\item \textbf{A manufactured phase transition} (\S\ref{sec:manufactured}). The construction
  exhibits a sharp absorbing-state transition, measurable to three decimal places, that is a
  property of the probe. We identify the mechanism (an attracting fixed point of the argmax map),
  delimit it (it occupies radius $r \in \{1,2\}$ only), and give the control that behaves as the
  mechanism predicts (a masked-LM construction, whose map has no such fixed point, shows no
  transition).
\item \textbf{The discriminator} (\S\ref{sec:discriminator}). Construction-determined and
  model-determined readings are separated by a test, not by intuition: vary one factor with the
  other held fixed. We tabulate five such manipulations. Varying the construction moves the
  instrument and varying the model across families does not, which is what makes the manufactured
  transition attributable to the probe; varying the training checkpoint does move it, which is what
  makes the developmental transition attributable to the model. The fifth manipulation --- varying
  family \emph{and} construction together --- separates the readouts from each other: $\lca$ does
  not survive it and the attractor share does.
\item \textbf{Estimator gating} (\S\ref{sec:gating}). Exponents measured on black-box LM dynamics
  must be gated at the measurement's own geometry, or they return confident wrong answers. We
  report four retracted verdicts, each caught by a known-answer system rather than by review, and
  ship the guards as a package.
\end{enumerate}

\paragraph{What this paper is not.} It is not an interpretability result. We do not claim $\lca$
localises a mechanism, and \S\ref{sec:limits} records the attempts that failed to make it do so.
The contribution is measurement: what an iterated self-feeding probe reads, what it does not, and
how to tell the difference before building on the answer.

\section{The construction}
\label{sec:construction}

Let $\mathcal{M}$ be a language model and $V$ its vocabulary. The state is a ring
$x \in V^{N}$ of $N$ token cells with periodic boundaries. One \emph{sweep} visits every site once
in a random order and resamples it in place from the model's conditional given its window:
\begin{equation}
x_i \;\sim\; p_r\!\left(\,\cdot \mid x_{i-r},\dots,x_{i-1},x_{i+1},\dots,x_{i+r}\right)
  \Big/ T ,
\label{eq:update}
\end{equation}
where $r$ is the radius, $T$ the temperature, and the centre token is masked from its own window.

\paragraph{Why these choices, and what they cost.} Each is a deliberate departure from how the model
is normally run, and stating them plainly is what allows \S\ref{sec:discriminator} to ask which
readings survive them. The ring is \emph{periodic}, so no site is privileged by being at a boundary
and the light cone is not confounded with an edge effect. Resampling is \emph{in place} rather than
appended, which is what closes the loop --- and it is also the single largest departure from
deployment, because free autoregressive generation never revisits a token it has emitted. The window
is \emph{symmetric with the centre masked}, which is native to a masked language model and is imposed
on an autoregressive one; that asymmetry is why we run both, and the masked-LM construction serves
throughout as the control. Sites are visited in a random order per sweep, and the order is drawn
\emph{per replica} rather than per batch, because sharing it across a batch correlates the replicas
and shrinks error bars by a factor they have not earned (\S\ref{sec:gating}).

\paragraph{The two-token window is genuinely two tokens.} A radius is a claim about geometry, and
it is cheap to check rather than assert: if the nearest token carried the window by itself, the
construction would be an $r=1$ chain wearing an $r=2$ label. On the autoregressive (causal,
left-window) construction we measure the exact coupled-draw disagreement produced by flipping each
window position separately --- the probability that twins sharing a uniform stream draw differently
when position $i-2$ changes, against the same for $i-1$. Averaged over six developmental checkpoints,
on the states the ring actually occupies, the far position contributes $0.579$ against the near
position's $0.820$, a ratio of $0.698$; at $r=3$ the third-back token still contributes $0.535$
against the nearest token's $0.704$. Influence decays with distance without the window collapsing
onto one position. The check earns its place because it can fail: restricting the same construction
to a small token sub-alphabet drops the far position to $0.061$ against $0.801$, the branching ratio
falls below one, and damage walks without growing.

The cause is not the restriction itself, and we say so because our own first reading was wrong. The
three sub-alphabets we tried are all semantically coherent sets, and the collapse tracks that
coherence rather than the alphabet's size: holding size and radius fixed and varying only how the
tokens are \emph{selected} moves the far position's contribution by up to $0.588$, with every
semantically chosen alphabet falling below a branching ratio of one and every randomly chosen one
sitting at or above it. Nor is the loss permanent --- widening the window recovers it, with all nine
arms reaching criticality by $r \le 6$. What the sub-alphabet lattice demonstrates is that this
construction \emph{can} be driven subcritical, which is what makes the full-vocabulary measurement a
check rather than a formality; it does not show that small alphabets are subcritical as such.

\paragraph{This is a probe, not a model of deployment, and we can say so quantitatively.} Injecting a
token error into real autoregressive generation and continuing produces no absorption at all:
$P_{\text{persist}} = 1.000$ on \texttt{pythia-70m}, \texttt{-160m} and \texttt{-410m} (32 trials
each), because free generation never resamples the damaged token. The dynamics studied here exist only when the loop is closed by resampling in place.
Any reading taken from them is therefore a statement about the closed loop first, and about the
model only if it survives \S\ref{sec:discriminator}.

\paragraph{Damage spreading under common random numbers.} We run two rings, $x$ and $\tilde{x}$,
identical except that one site is flipped at $t=0$, and advance both with the \emph{same} stream of
uniform variates and the \emph{same} visit order. Writing $d_t$ for the number of differing sites
after $t$ sweeps, the token-space Lyapunov exponent $\lca$ is the exponential growth rate of $d_t$
over the sweeps before saturation, and $\dnorm$ is the saturating $d_t$ normalised by the
independent-noise floor $D_0$ obtained when the two rings are driven by \emph{independent} uniforms.
The two quantities answer different questions --- how fast damage grows, and how much survives ---
and \S\ref{sec:gating} explains why they must be filtered differently.

\begin{figure}[t]
\centering
\includegraphics[width=\linewidth]{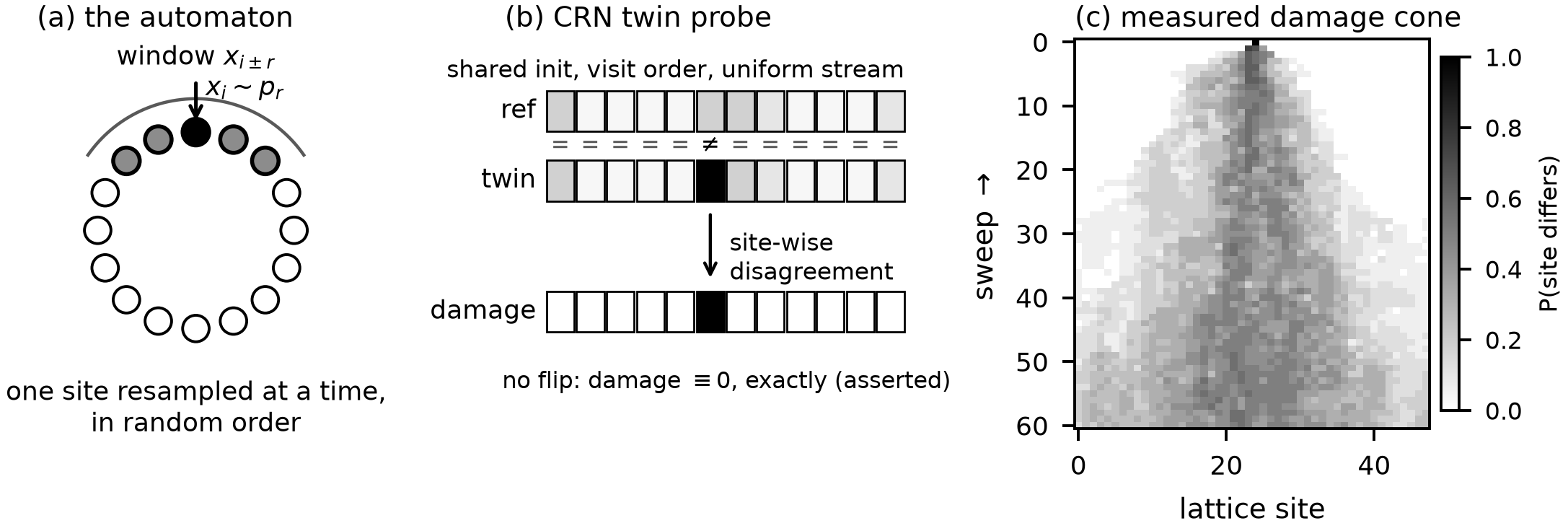}
\caption{The construction. A ring of token cells is resampled in place by the model's own windowed
conditional; two copies differing in one token are advanced under common random numbers, and the
growth of their disagreement gives $\lca$. Undamaged twins diverge by exactly zero.}
\label{fig:instrument}
\end{figure}

\paragraph{The exact-zero null.} Two rings with identical windows, handed the same uniform, draw
the same token: inverse-CDF sampling against a shared uniform makes agreement exact, not
approximate. Undamaged twins therefore diverge by \emph{exactly} zero, on every backend, and this
is asserted rather than assumed. It is the property that makes damage attributable to the injected
flip and to nothing else, and every claim in this paper rests on it.

\section{Validation by reproduction}
\label{sec:validation}

An instrument that has only ever been pointed at an object with no known answer cannot be
distinguished from a plausible-looking implementation that is wrong. We therefore calibrate by
\emph{reproduction}, on systems whose answers are established independently, and in an order that
climbs from the strictest test to the loosest.

\paragraph{Rung 1: an identity, not a correlation.} On the Domany--Kinzel probabilistic cellular
automaton the damage field is provably the automaton itself, run on a derived rule. This yields a
prediction with \emph{no error bar attached}: on the $p_2 = 0$ line the CRN damage field is itself a
Domany--Kinzel automaton at the same $p_1$, so ours must equal an independently predicted one cell
for cell. It does --- \textbf{0 mismatching cells} at $p_1 \in \{0.2, 0.5, 0.75, 0.8087, 0.95,
1.0\}$, on a ring of 4096 over 1500 steps, three seeds each --- while an off-line control at
$(0.6, 0.5)$ yields 16 mismatches, so the test is not vacuous. The identity runs \emph{through} the
same loop that produces every language-model number here, so it verifies the window indexing, the
shared-uniform consumption order, the inverse-CDF sampling and the synchronous update at once. No
other rung does that; the rest agree only to within a fitted constant.

\paragraph{Rung 2: a known ordering, and the part of it the instrument does not recover.} Across
elementary cellular automata, ignition probability separates the ordered rules from the rest
decisively (Cohen's $d = 3.03$, $p < 10^{-3}$). It does \emph{not} separate edge-of-chaos from
chaotic ($p = 0.47$). We report the rung as what it is --- a recovered ordered/non-ordered boundary,
not a recovered three-way classification --- because a rung that is claimed to do more than it does
stops being a calibration.

\paragraph{Rung 3: known distributions.} Run on systems with known transition matrices, the
attractor census recovers them.

\begin{figure}[t]
\centering
\includegraphics[width=\linewidth]{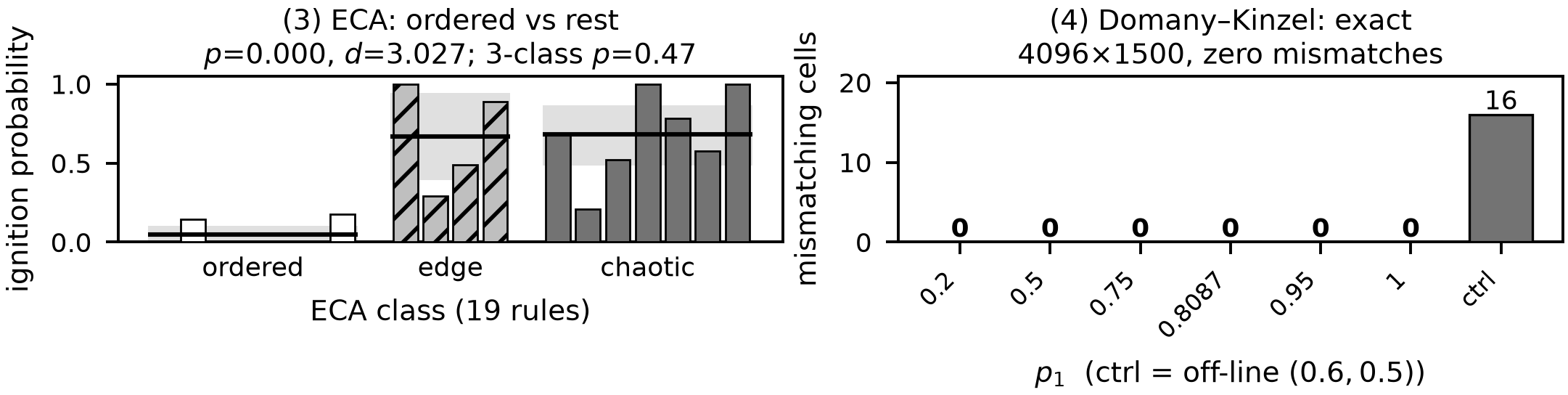}
\caption{Validation by reproduction. The instrument is calibrated against systems whose answers are
established independently before it is pointed at a language model, strictest rung first: the
Domany--Kinzel identity admits no error bar, so a merely-close implementation fails it.}
\label{fig:ladder}
\end{figure}

Only after all three rungs does the instrument get pointed at a language model. This ordering is
not ceremonial: \S\ref{sec:gating} reports four occasions on which a rung caught an estimator that
had already produced a confident number.
\section{A phase transition that belongs to the probe}
\label{sec:manufactured}

Iterated at low temperature, the construction shows a sharp absorbing-state transition. Both the
survival exponent $\delta$ and the density exponent $\theta$ reach their directed-percolation values
at a common critical temperature $T_c \in [0.4343, 0.4391]$, and the estimator was gated on
Domany--Kinzel \emph{before} the language-model numbers were read.

\begin{figure}[t]
\centering
\includegraphics[width=\linewidth]{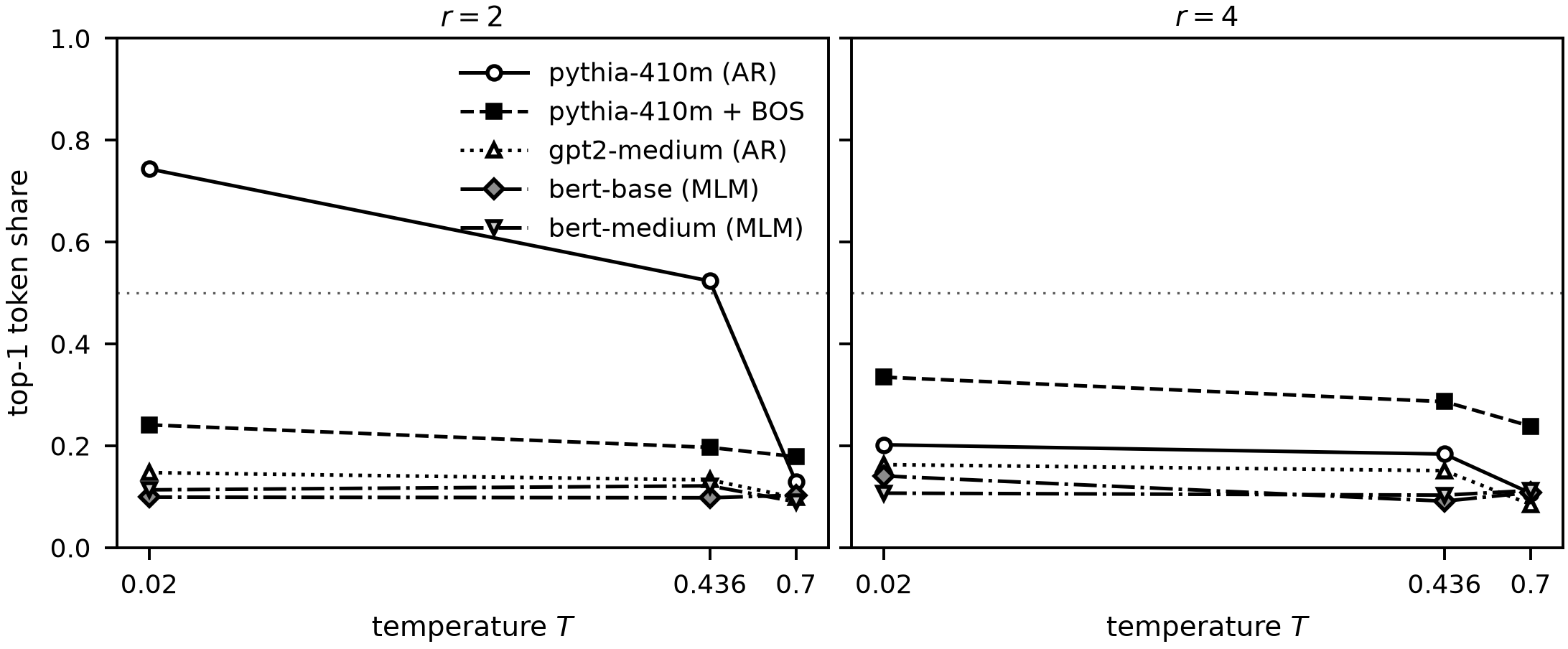}
\caption{The transition belongs to the probe. \textbf{Left ($r=2$):} under the autoregressive
construction \texttt{pythia-410m} collapses onto a single token as $T$ falls, reaching a top-1 share
of $0.744$ at $T=0.02$, and crosses half the ring at the measured $T_c$. Three controls do not.
Prepending one beginning-of-sequence token to the same model drops it to $0.241$ --- the map's
domain changes, not its parameters. \texttt{gpt2-medium}, whose argmax map has no attracting fixed
point, sits at $0.147$. Two masked-LM models, whose \emph{native} task is this update, stay at
$0.099$ and $0.113$ at every temperature. \textbf{Right ($r=4$):} outside the degenerate radius all
five arms look alike, which is the boundary. Monochrome by marker and dash throughout.}
\label{fig:manufactured}
\end{figure}

It is a property of the probe. Four facts establish this, and the fourth is the mechanism, and
Figure~\ref{fig:manufactured} carries all four on one grid.

\paragraph{The frozen phase is a single-token collapse.} At $T = 0.02$, 81 of 96 sites hold the
newline token, and the measured $T_c$ sits at the point where newline occupies 52\% of the ring.
The ``ordered phase'' is one token eating the lattice.

\paragraph{It is not the corpus, the architecture, or the scale.} The effect is refuted from both
directions across 19 models. Granite's dense and mixture-of-experts members agree within two points
while differing by $2\times$ in width, $1.7\times$ in depth and $16\times$ in feed-forward size, and
in routing versus none. Scale is eliminated across a $70\times$ Pythia ladder and a $12\times$ GPT-2
ladder whose ranges never overlap.

\paragraph{The mechanism is an attracting fixed point of the argmax map.} At $T = 0.02$ the update
is essentially deterministic, so the dynamics are governed by the map $x \mapsto \arg\max p_r$.
For \texttt{pythia-410m} this map sends 18 of 24 random starts to the newline token --- a genuine
fixed point. For \texttt{gpt2-medium} it has no such point and wanders to 11 distinct endpoints.
Prepending a single beginning-of-sequence token moves the frozen fraction from 74.4\% to 24.1\%,
because it changes the map's domain rather than its parameters. This is not a data-sparsity effect:
rare contexts are no closer to the fallback token than common ones.

\paragraph{The boundary is sharp and narrow.} Family-distinguishing degeneracy occupies radius
$r \in \{1,2\}$ only; moving from $r=2$ to $r=3$ drops top-1 agreement by 52 points. A rebound at
large radius appears in the control as well, so it is a generic long-context effect and is excluded.

\paragraph{The control behaves as the mechanism predicts.} A masked-LM construction, whose argmax
map has no attracting fixed point, shows no transition: surviving damage never falls below 0.547
down to $T = 0.02$ across two models. This was pre-registered as a good null --- no absorbing state,
therefore no absorbing-state transition --- so there is no competing ``but the clean construction
has a real one'' left to explain.

Taken together these make the transition a \emph{claim} rather than a curiosity: it has a mechanism,
a boundary, and a control that behaves as the mechanism says it should. It is also, we emphasise,
a real and reproducible measurement --- of the probe.

\section{The discriminator: which readings are the model?}
\label{sec:discriminator}

The manufactured transition of \S\ref{sec:manufactured} raises the obvious worry: if a quantity
this sharp belongs to the probe, does the instrument read the model at all? It does, and the way to
establish it is not argument but a second manipulation. A reading that is construction-determined
moves when the construction changes and stays put when the model changes; a model-determined
reading does the reverse.

\begin{table}[t]
\centering
\begin{tabular}{llll}
\toprule
Manipulation & Construction & Model & Instrument responds? \\
\midrule
Change the automaton (\S\ref{sec:manufactured}) & varied & fixed & yes --- transition disappears \\
Change family or scale                          & fixed  & varied & \textbf{no} --- 19 models, $70\times$ \\
Change family, \emph{construction varied}       & varied & varied & \textbf{no} for $\lca$; \textbf{yes} for the share \\
Change the training checkpoint                  & fixed  & varied & yes --- $\lca$ crosses zero \\
Ablate internal components                      & fixed  & varied & yes --- ignition moves by up to $0.33$ \\
\bottomrule
\end{tabular}
\caption{The discriminator. A reading determined by the construction moves when the construction
changes and not when the model changes; a model-determined reading does the reverse. Row~2 is what
makes the manufactured transition attributable to the probe. Rows~3 and~4 are what establish that
the instrument reads the model at all.}
\label{tab:discriminator}
\end{table}

Table~\ref{tab:discriminator} sets out the four manipulations. The pattern, not any single row, is
the argument: an instrument that responded to everything would be measuring noise, and one that
responded to nothing would be measuring the construction alone.

\subsection{Construction-determined readings}
Two readings that look like measurements are fixed by the geometry of the probe.

The \emph{damage light cone} is kinematic. A flipped site can influence only the $r$ sites whose
window contains it, so the cone's extent is set by the update window rather than by the model, and
its slope is a geometric fact reported in the units of a dynamical quantity --- the lattice analogue
of a Lieb--Robinson bound \citep{lieb1972finite}, where the propagation limit is a property of the
interaction structure and not of the state.

We state the bound carefully, because the natural sharper version is false here. Updating is
\emph{asynchronous in random order}, so within a single sweep a site damaged early can pass damage to
its right neighbour, which is then itself visited: the reach inside one sweep is bounded by the visit
order, not by $r$. Measured directly, the front reaches offset 24 by sweep 8 where $r\cdot t = 16$.
The cone is therefore kinematic in the sense that matters --- it carries no model information --- but
``at most $r$ sites per sweep'' is a synchronous bound and does not hold for this construction.
The cone's \emph{shape} likewise adds no resolution: the width of its front has span exactly zero
across every checkpoint and seed at two ring sizes --- a null that survives a fourfold change in
ring size, a tenfold change in the resolvable window and a thirtyfold change in how densely the cone
is filled.

The \emph{radius scaling} $\lca(r)$ is model-invariant. If it read the model it would differ between
models; it does not. This is the cleanest available demonstration that a quantity can be stable,
reproducible, and precisely measurable while carrying no information about the system one believes
oneself to be measuring.

\subsection{Model-determined readings}
Hold the construction fixed and walk the model through training. At $N = 48$ on Pythia-410m,
$\lca$ reads $-0.0185$ at step 256, $+0.0679$ at step 512, $+0.1923$ at step 1000, and settles on a
plateau of $+0.1558$, $+0.1699$, $+0.1792$ at steps 2000, 8000 and 143000. Before the crossing,
seeds disagree about the \emph{sign}; after it, every one of 48 runs is positive, at two lattice
sizes, with all four members of a pre-registered family surviving Benjamini--Hochberg correction.
The automaton is byte-identical at every checkpoint, so the movement is the model.

\begin{figure}[t]
\centering
\includegraphics[width=\linewidth]{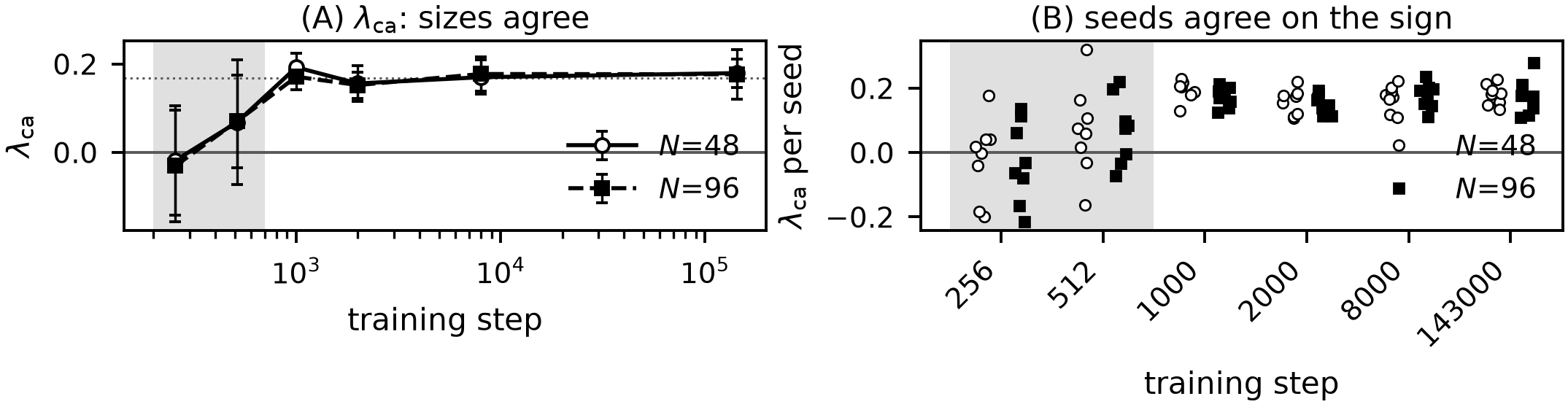}
\caption{A model-determined reading. With the automaton byte-identical at every checkpoint, $\lca$
crosses zero at a reproducible point in training: before it, seeds disagree about the sign; after
it, all 48 runs are positive at two lattice sizes. Contrast Table~\ref{tab:discriminator} row~2,
where varying the model across 19 architectures and a $70\times$ scale range moves nothing.}
\label{fig:developmental}
\end{figure}

\paragraph{The scope of this reading is within-model, and we state it here rather than in the
limits.} Holding the construction fixed and walking one model through training is what the paragraph
above does, and the movement is large. Varying the \emph{model} instead does not move $\lca$ usefully:
across ten models spanning six families and four architecture classes, its cross-model spread is
$0.051$, against a range of $0.122$ to $0.804$ produced by varying radius and temperature alone. The
ordering that spread implies is not reproducible --- seed stability $0.030$ --- and reshuffles between
temperatures. Most directly, $\lca$ does not see the architectural contrast of
\S\ref{sec:discriminator}: RWKV, which has no attractor at all, sits mid-pack. $\lca$ is a
developmental quantity, not a model-comparison one, and none of this touches the curve above, whose
range is roughly seven times the cross-model spread.

\paragraph{The attractor share is the model-determined reading that survives the construction.}
Where $\lca$ fails, the attractor share --- the dominant token's occupancy of the settled ring ---
passes every corresponding check. Across the same ten models and six constructions (two radii, three
temperatures) its across-model spread exceeds its across-seed spread on $6$ of $6$ constructions
against $\lca$'s $2$ of $4$; its model ordering is seed-stable at $0.848$ against $0.030$; and that
ordering agrees across constructions at $\rho = +0.752$, with two further readouts --- distinct-token
count and adjacent-pair repetition --- clearing the same threshold independently at $+0.737$ and
$+0.654$. Its cross-model spread is $0.923$. Controlling for corpus, it recovers the architectural
effect $\lca$ misses: RWKV sits $0.769$ below the Pile-trained attention models. The instrument's
transferring results are built on this quantity, and it is the one that is model-attributable.

The obvious deflation --- that $\lca$ is a repackaged loss --- does not survive measurement. Across
the trained regime $\lca$ moves from $0.18433$ to $0.18738$ while bits-per-byte improves from
$2.3229$ to $0.8895$, a $2.6\times$ change in model quality against a change in $\lca$ that sits
inside the seed spread. Whatever the exponent tracks, it is not simply how good the model is.

\subsection{Ablation response}
\label{sec:ablation}
Hold the construction fixed and remove part of the model. Ablating the early attention block leaves
the lattice nearly frozen --- ignition falls from $0.977$ to $0.181$ at the final checkpoint --- and
adding one \emph{further} attention ablation moves it again, by up to $0.33$. That the instrument
responds at all to an internal manipulation, with every construction parameter unchanged, is what
row~4 of Table~\ref{tab:discriminator} records.

\emph{How} it responds is worth stating carefully, because the obvious reading of a single
checkpoint is wrong. At the final checkpoint the further ablation \emph{raises} ignition, and five
separate layers do so at Bonferroni-corrected significance, which invites the conclusion that
removing more of a network makes its dynamics livelier. Measured across five post-crossing
checkpoints, that conclusion does not survive. Regressing compound ignition on reference ignition
gives a slope of $+0.568\,[+0.461, +0.674]$ for one layer and $+0.724\,[+0.618, +0.830]$ for
another: both intervals exclude $0$, so the compound arm is not sitting at a fixed level, and both
exclude $1$, so it is not tracking its reference with a constant offset either. The compound arm
varies less than what it follows (standard deviations $0.229$ and $0.216$ against $0.304$).

The behaviour is \textbf{partial regression toward an intermediate value}: an added ablation moves
ignition part of the way toward a middle, so the \emph{sign} of the change depends on where the
reference already sits. At the final checkpoint the reference is frozen near the bottom of its
range and everything rises; at an earlier checkpoint, where the reference sits at $0.581$, the same
two layers move in opposite directions --- one falls to $0.397$ while the other rises to $0.781$.
No property of those particular layers explains that, and no account in which the effect is
anti-monotone survives it.

We report this as a description of the ablation response rather than as a mechanism. It is what row
4 needs --- the instrument reads the model --- and it is deliberately less than the single-checkpoint
measurement appeared to offer.

\section{Gating estimators at their own geometry}
\label{sec:gating}

An exponent measured on black-box language-model dynamics will return a number whatever you do.
The number is not the finding; whether the estimator had room to produce a different one is. Four
verdicts in this project were retracted, and none was caught by review --- each was caught by a
system whose answer was already known.

\paragraph{Gate the estimator at the measurement's own geometry.} A directed-percolation
calibration performed at $N=512$ over 200 sweeps licenses nothing at $N=96$ over 40. The first
retraction was a calibration run at a geometry the measurement never used.

\paragraph{State what the independent unit is, and test it.} Anything drawn once per batch makes
replicas correlated. One visit order, shared across a batch, decided the whole result and inverted
the verdict; pooling correlated replicas shrank the error bars by roughly $8\times$ for a factor
they had not earned.

\paragraph{A cost function that can shrink its own comparison window is unbounded.} Scan far wider
than plausible, and \emph{reject} a minimum that lands on the edge of the scan.

\paragraph{Show the test can discriminate before quoting it.} A transverse-Lyapunov test was
demonstrated, on a system with a known answer, to be unable to separate the hypotheses at all ---
so its reading on the model was uninterpretable in either direction rather than weak evidence.

\paragraph{A mean is a claim about the axis it averages over.} A scalar summary is quotable only if
the axis it reduces varies less than the effect the summary is being used to describe; otherwise the
number characterises the reduction rather than the quantity. That is directly checkable --- compare
the spread \emph{within} the reduced axis against the movement of the mean \emph{across} conditions
--- and applied to every per-component dataset in this project it flags the \emph{ensemble} axis
twice while clearing the window-position axis, recovering by construction a distinction we had
previously found by hand. We report it here rather than as advice because it retracted the finding
that motivated it. A claim that one of our own mean-field inputs had been computed on the wrong
summary failed the regression test we built for it from that finding's own data; the two summaries
turned out to be algebraically identical, the claim was withdrawn the same day, and the measurement
it had been attached to stood unchanged.

\paragraph{Undefined is not zero.} $\lca$ is emitted for runs in which damage never ignited, where
it is undefined; the same runs carry a $\dnorm$ of exactly zero, which is a true measurement. The
two quantities therefore need \emph{different} filters --- $\lca$ over ignited runs only, $\dnorm$
over all of them --- and collapsing that asymmetry biases the metric that was not broken. The
mechanism is prosaic: roughly one visit order in three heals a single-site seed before it can
propagate.

\begin{table}[t]
\centering
\small
\begin{tabular}{p{0.28\linewidth}p{0.34\linewidth}p{0.28\linewidth}}
\toprule
Defect & What it produced & Guard \\
\midrule
Calibration run at a geometry the measurement never used
  & a directed-percolation verdict licensed at $N{=}512/200$ and read at $N{=}96/40$
  & gate the estimator at the measurement's own geometry \\
One draw shared across a batch
  & correlated replicas, error bars $\sim 8\times$ too small, verdict inverted
  & state the independent unit and test it \\
Cost function able to shrink its own comparison window
  & an unbounded minimum, landing on the edge of the scan
  & scan wider than plausible; reject edge minima \\
Test never shown able to discriminate
  & a reading uninterpretable in either direction, quoted as weak evidence
  & demonstrate discrimination on a known answer first \\
Undefined treated as zero
  & $\lca$ averaged over runs where damage never ignited
  & different filters for $\lca$ and $\dnorm$ (\S\ref{sec:construction}) \\
\midrule
\multicolumn{3}{l}{\emph{After all five guards existed:}} \\
Wrong error bar for the statistic
  & standard error of a difference of four centres taken as their \emph{mean}, understating it
    $\sim 2\times$; a positive recorded and withdrawn the same day
  & derive the error bar for the statistic at hand, not for the quantity it is built from \\
\bottomrule
\end{tabular}
\caption{Four retracted verdicts, one further defect of the same class, and one that arrived after
all the guards were in place. None was caught by review: each was caught by a system whose answer
was already known, or by writing the follow-up experiment. Every row is the same underlying error
--- a statistically-shaped criterion applied to a quantity with no room to vary.}
\label{tab:retractions}
\end{table}

\paragraph{One defect class.} All of these are the same error: \emph{a statistically-shaped criterion
applied to a quantity with no room to vary}. A correlation whose predictor is saturated, a ratio
whose denominator is noise, a directional hypothesis tested with an absolute value --- each returns
a confident number from a comparison that could not have come out otherwise. The class itself is
old: psychometrics has known its data side for a century as restriction of range and ceiling
effects \citep{pearson1903,sackett2000range}, and its classical response is to \emph{correct} the
attenuated estimate. The guards here take the other branch --- refuse the verdict --- and the same
remedy was arrived at independently and concurrently from the survey-methodology side: an audit of
question-order effects in an instruction-tuned model found 17 of 18 item pairs saturated under a
forced-binary next-token readout and recommends a pre-specified saturation diagnostic as a
standard health check for any study that reads next-token probabilities as response distributions
\citep{kang2026qq}. We ship the guards as
an MIT-licensed package: a dynamic-range check on the target, a noise gate before any ratio, a
directional test where the hypothesis is directional, an explicit \textsc{not decidable} branch,
the same range check applied to the \emph{predictor}, and a distinct-context floor on the
estimator's own input.

\paragraph{The class survives its own countermeasures.} The most instructive instance came after all
six guards existed. A single line --- the standard error for a difference of four independently
measured centres --- was written three ways over one analysis. Dividing the pooled spread by
$\sqrt{8}$ used a seed count that had gone stale after the design was extended, overstating the
noise and returning \textsc{not decidable}. Replacing it with the \emph{mean} per-arm standard error
fixed the stale count but used the wrong statistic for a difference of four quantities, understating
the uncertainty by about a factor of two and producing a positive result. Only the quadrature sum is
correct, and it returns \textsc{not decidable} again. The middle version is the one that produced a
finding; it was recorded and withdrawn the same day. It was caught not by the guards but by writing
the \emph{confirmatory} experiment, whose floor was derived from first principles for the statistic
at hand rather than inherited from the sweep. A pre-registration, a power calculation and a fixed
stopping rule all passed it through.

\paragraph{And one instance the guards were structurally unable to see.} The guards above inspect
the \emph{data}: whether a target has range, whether a denominator is noise, whether a predictor is
saturated. The sharpest instance of the class was in none of those places. Every correlation in the
project ranked with \verb|argsort(argsort(x))|, which is correct only when all values are distinct
--- \verb|argsort| breaks ties by input position, so a constant vector receives strictly increasing
ranks. It fired in production: a shape scalar whose twenty-four measured values were all exactly
$0.000$ was reported as correlating with the growth rate at $\rho = +0.829$, $p = 0.058$. The
reported number was the correlation between the growth rate and \emph{the order the checkpoints
happened to be listed in}. The data was honest --- the scalar was a correctly measured constant ---
so no data-level gate could have flagged it; the defect was in the correlation function. Fifteen
scripts carried the idiom. Re-running all of them with tie-aware ranking on identical stored inputs
moved five results and changed no conclusion, and the externally predictive one was never exposed.
The guard that closes it is not another data check but a primitive returning \textsc{nan} on a
zero-variance input, plus a test that greps the repository so the idiom cannot return by copy-paste.

\section{Limits}
\label{sec:limits}

\paragraph{The developmental transition is single-family.} It is measured on Pythia. Endpoints
replicate in two non-Pythia families, but no public non-Pythia family publishes a checkpoint inside
the window where the crossing occurs, so the transition's \emph{shape} is unobservable outside
Pythia by anyone, not merely by us.

\paragraph{Cross-family, loss does not organise it better than tokens.} Matching families on
bits-per-byte gives an across-family spread of $0.0588$ against $0.0318$ when matching on token
count --- a difference of $1.37\times$ the seed floor, under our $2\times$ gate. That comparison is
\textsc{not decidable}: underpowered rather than null, and the fix is finer checkpoint spacing,
which does not exist to be had. A separate question on the same grid \emph{is} decisive: the
matched-bpb spread alone stands at $3\times$ the floor, so the families do not collapse onto one
curve, and $\lca$ is not a function of model quality.

\paragraph{The explanandum is not internal: $\lca$ is largely fixed by the state the model drives
the lattice into.} Four routes failed to attach $\lca$ to a named internal mechanism --- co-timing
against context-use onset, ablation of component groups, induction-head formation (excluded by
arithmetic, since formation lands one to two orders of magnitude away from the window), and a
compensator-identification test whose single positive was withdrawn when its standard error was
corrected. A fifth succeeded by changing level. The number of distinct tokens in the settled ring
rank-correlates with $\lca$ at $\rho = 0.771$ (bootstrap 95\% CI $[0.714, 0.829]$ over eight seeds
per checkpoint), and the relation is not a shared trend with training time: holding the weights
fixed and varying temperature instead, cells three orders of magnitude apart in training land
together when their diversity matches --- $T{=}0.9$ at step~256 and $T{=}0.5$ at step~143000 give
diversity $21.6$ versus $26.8$ and $\lca$ $+0.187$ versus $+0.183$. Pooling both checkpoints onto one
diversity--$\lca$ curve costs $0.009$ in residual against a $0.046$ seed floor.

\paragraph{This retrodicts the four failures.} They searched for an internal cause of a quantity
fixed by the state, and the sharpest of them reads differently in that light: no single attention
layer moves $\lca$ beyond seed scatter, eight together move it by $+0.345$, and the twenty-four
singles sum to $-0.224$ --- the wrong sign. That is not a mystery about localisation; it is what a
collective property of the settled state looks like under ablation. The reduction is statistical
rather than an identity --- roughly forty per cent of $\lca$'s variance is not diversity --- and it
names no circuit and no training event. What it removes is the expectation that one exists at the
level the four routes searched.

\paragraph{And the reduction does not extend to the predictive result.} $\Tstar$ is derived from the
same settled ring, which raises the possibility that it is the reduction in different clothing. It is
not: diversity at a \emph{fixed} temperature predicts greedy degeneration at $|\rho| \le 0.11$ across
four temperatures on twenty-six models, every $p > 0.59$, while $\Tstar$ on the same target and the
same models reaches $\rho = 0.547$. The predictive content lies in where the diversity curve crosses
a threshold as temperature varies, not in diversity at any point on it --- so $\lca$ inherits the
state's lack of external predictive power, and $\Tstar$ does not.

\paragraph{That asymmetry is target-specific, and reverses on the one other target we tested.} The
reading above --- that what transfers is a \emph{response} (how the settled state dissolves under
temperature) rather than a \emph{level} (the state itself) --- holds for greedy degeneration, where it
was derived. Asked instead which readout predicts \emph{instruction-following} failures, the ordering
flips: the attractor share, a level, is selective for compliance at $+0.53$ ($p = 0.004$, $n = 10$),
surviving controls for both model size and general capability, while $\Tstar$ is not
($+0.17$ against a verified $+0.34$ detection floor at $n = 6$, so an effect of the share's size
would have been found). Repetition rate is not selective either, so the share's result is not
mediated by degeneration. We therefore do not claim a general recipe. Which of the two transfers
depends on what is being predicted, and one confirmed instance in each direction does not establish
why.

\paragraph{The dynamics are not reducible to the obvious theory of them.} Given the model's own
exactly measured single-token sensitivity, annealed mean field does not predict the exponent across
33 ablation arms --- a null with adequate power, where the predictor spans nearly twice the target's
range. Whatever sets $\lca$, it is not captured by the mean-field ledger.

\paragraph{Scope.} Greedy decoding where decoding matters; one radius for the ablation work; one
architecture family for the component manipulations. These bound the claims rather than qualify
them: the discriminator of \S\ref{sec:discriminator} is a method, and the specific readings it
sorts are the ones we measured.

\section{What this opens}
\label{sec:future}

Three things follow directly, and the first is the one we would do next.

\paragraph{The coupling is a common mode, and that is now measured rather than assumed.} Every
relative reading in this paper is taken under one coupling, so it is fair to ask whether the choice
is doing the work. Running the developmental checkpoints under \emph{both} the monotone coupling
used throughout and a maximal coupling, on the same rings and the same models, the ordering of the
checkpoints is identical and the offset between the two is uniform within the seed floor (offsets
of $-0.039$, $-0.087$, $-0.036$ against a floor of $0.036$). Maximal reads lower everywhere, which
is the expected direction: it maximises agreement between the twins and therefore minimises damage.
So the readings in \S\ref{sec:discriminator} are properties of the model rather than of the
coupling, on this geometry.

Two things this does \emph{not} settle, and they are the natural next measurements. The result
covers one radius, one temperature and one family, so coupling-invariance in general is untested.
And it is a statement about \emph{relative} readings only: absolute damage does differ between
couplings, so any absolute figure must name the coupling it was taken under. What remains genuinely
open is the comparison in the other direction --- mixing time is a functional of the marginal chain
and is therefore coupling-invariant by construction, while damage velocity is a functional of the
coupled chain, so measuring both on one substrate would say how much of the dynamics a coupling
choice can reach at all.

\paragraph{Apply the discriminator to the practices that motivated it.} Self-consistency, iterated
refinement and agentic loops are all self-feeding systems whose readings nobody separates into
construction and model. The test of \S\ref{sec:discriminator} does not depend on our construction:
it needs only two manipulations, one that varies the loop with the model fixed and one that varies
the model with the loop fixed. Whether the quantities practitioners already read off those systems
--- agreement rates, convergence speed, self-correction success --- survive that test is an open and
answerable question, and the manufactured transition of \S\ref{sec:manufactured} is the reason to
ask it before building on them.

\paragraph{A construction axis rather than a construction.} Radius, temperature, visit scheme and
masking are a parameter family, not a fixed choice, and they dial how much of the system is loop.
Sweeping them turns the discriminator from a two-point test into a gradient: readings can be sorted
by \emph{how fast} they decay as the construction is loosened, which would separate the sharply
kinematic from the merely construction-sensitive rather than treating both as one category.

\section{Related work}
\label{sec:related}

We ran an explicit prior-art check and report both what it found taken and what it found open,
because claiming novelty we do not have is the fastest way to lose a reader who knows this
literature.

\paragraph{Taken.} Detecting that an API is serving a distorted model is established: framed as a
two-sample problem, a test on the maximum mean discrepancy between an API's outputs and a reference
distribution reaches a median 77.4\% power against a range of distortions from roughly ten samples
per prompt, and finds 11 of 31 surveyed Llama endpoints deviating from the released reference
weights \citep{modelequality}. Text-only identification is stronger still: visible-string features
over random-string probes reach a verification AUROC of $0.99$ on a same-family ladder and separate
every endpoint of a commercial gateway within a handful of calls \citep{iris}. Quantization
detection is likewise established, and the tokenizer-merge mechanism we observed in API-mediated
probing is substantially anticipated. We claim none of these.

\paragraph{Iterated generation is not new; this measurement of it is.} Feeding a model its own
output has been studied as a dynamical system before. Multi-turn transmission chains exhibit
cultural attractors \citep{telephone}, successive paraphrasing settles into attractor cycles
\citep{paraphrase2cycle}, and the output distribution of a language model shows phase transitions
under temperature \citep{phase_detect}. We do not claim the observation that iteration has
attractors.

What we add is a \emph{measurement} of the iterated system rather than a description of where it
settles: damage spreading under common random numbers, with an exact-zero null, giving a Lyapunov
exponent and a damping length --- and, in the model-identification literature specifically, every
published feature set performs single-shot scoring of supplied text rather than measuring dynamics
at all. The contribution that does not follow from prior work is the discriminator of
\S\ref{sec:discriminator}: the attractor literature reports what the iterated system does, and does
not separate which of those readings is a property of the iteration and which is a property of the
model. That separation is what this paper supplies, and \S\ref{sec:manufactured} is what happens
without it.

\paragraph{Inherited machinery.} Damage spreading and the Domany--Kinzel automaton come from the
statistical-physics literature on probabilistic cellular automata \citep{domany1984,
bagnoli1992damage}, and we use them as calibration targets rather than as objects of study. We cite
the self-repair literature \citep{hydra,selfrepair} for terminology only: we avoid the word
``repair'' for our own repair-length quantity because of that collision, and we do not claim to
have measured self-repair.

\section{Conclusion}
\label{sec:conclusion}

Iterated self-feeding probes mix construction-determined and model-determined quantities in
readings that look alike. That is the finding, and the test in \S\ref{sec:discriminator} is what
separates them: vary one factor with the other held fixed, and see which readings move.

The worked example is a phase transition we measured to three decimal places, with directed
percolation exponents reached at a common critical temperature, that belongs to the probe. It has a
mechanism, a boundary and a control; it is entirely reproducible; and it says nothing about any
language model. The same instrument, on the same construction, also detects a reproducible
developmental transition that \emph{is} the model. Nothing distinguishes the two by inspection.

We resist the neat summary --- that the instrument measures itself --- because it is false and would
license the wrong conclusion. The instrument reads the model. It also reads the probe, in quantities
that carry the same units and the same apparent precision, and the cost of not checking which is
which is four months and a retracted phase transition.

The sharper statement is that \emph{which} of its quantities reads the model is itself something to
be measured rather than assumed, and the answer is not uniform across them. Under construction
variation the attractor share ranks models consistently and $\lca$ does not; $\lca$'s domain is one
model's trajectory through training, where its range is an order of magnitude larger than anything
model identity produces. Both are quantities of the same instrument, read off the same rings, in
the same units. An instrument can be valid for one comparison and empty for another, and no amount
of care in the measurement reveals which without varying the apparatus.

\subsubsection*{Reproducibility}
All code, per-run result files and figure scripts are released. Every number in this paper is
traceable to a results file, and each analysis stamps the SHA-256 of its own source together with
the hashes of every module it imports, so a figure cannot silently outlive the code that produced
it. The estimator guards are packaged separately under an MIT licence.
The repository is \url{https://github.com/nicoveraz/token-lattice-ca} and is archived at
\url{https://doi.org/10.5281/zenodo.21880472} (all versions): code under MIT, the prose and the findings ledger
under CC BY 4.0. The ledger retains retracted and amended findings in place rather than deleting
them, so a citation should carry the amendment with the finding.

\bibliographystyle{plainnat}
\bibliography{refs}

\end{document}